\documentclass{article}
\usepackage{arxiv}

\pdfoutput=1

\usepackage{amsmath}
\usepackage{graphicx,changepage}

\usepackage{framed}
\usepackage[hidelinks]{hyperref}
\usepackage{natbib}     

\usepackage{xcolor}
\usepackage{multicol}
\usepackage{multirow}
\usepackage{makecell}

\usepackage{url}

\title{Discovering and Explaining Driver Behaviour under HoS Regulations}

\usepackage{authblk}

\author[1]{Ignacio Vellido}
\author[1]{Juan Fdez-Olivares}
\author[1]{Ra{\'u}l P{\'e}rez}

\affil[1]{Department of Computer Science and Artificial Intelligence, University of Granada, Spain}

\affil[ ]{}

\affil[ ]{\textit{ignaciovellido@ugr.es, \{faro, fgr\}@decsai.ugr.es}}

\renewcommand{\shorttitle}{Discovering and Explaining Driver Behaviour under HoS Regulations}

\hypersetup{
pdftitle={Discovering and Explaining Driver Behaviour under HoS Regulations},
pdfsubject={q-bio.NC, q-bio.QM},
pdfauthor={I. Vellido, C. N{\'u}\~{n}ez-Molina, V. Nikolov and J. Fdez-Olivares},
pdfkeywords={Explainable AI, activity recognition, vehicular technology, hours of service, tachograph, infringement detection},
}

\begin{document}
\maketitle

\begin{abstract}
World wide transport authorities are imposing complex Hours of Service regulations to drivers, which constraint the amount of working, driving and resting time when delivering a service. As a consequence, transport companies are responsible not only of scheduling driving plans aligned with laws that define the legal behaviour of a driver, but also of monitoring and identifying as soon as possible problematic patterns that can incur in costs due to sanctions. Transport experts are frequently in charge of many drivers and lack time to analyse the vast amount of data recorded by the onboard sensors, and companies have grown accustomed to pay sanctions rather than predict and forestall wrongdoings. This paper exposes an application for summarising raw driver activity logs according to these regulations and for explaining driver behaviour in a human readable format. The system employs planning, constraint, and clustering techniques to extract and describe what the driver has been doing while identifying infractions and the activities that originate them. Furthermore, it groups drivers based on similar driving patterns. An experimentation in real world data indicates that recurring driving patterns can be clustered from short basic driving sequences to whole drivers working days.
\end{abstract}


\section{Introduction}

World wide transport authorities are imposing complex \textbf{H}ours \textbf{o}f \textbf{S}ervice (from now on, HoS) regulations to drivers \citep{meyer_european_2011,goel2013hours}, which constraint the amount of working, driving and resting time when delivering a service. As a consequence, transport companies are responsible not only of scheduling driving plans aligned with laws that define the legal behaviour of a driver, but also of monitoring and identifying as soon as possible problematic patterns that can incur in costs due to sanctions.

Fortunately, the widespread adoption of onboard IoT devices in vehicle fleets enables recording of the driver activities in event logs, but the large amount of data ingested makes difficult for transport experts to understand what happened and to make actions that forestall illegal behaviour. For this reason, an important technical challenge is to come up with easily interpretable descriptive models that help understand the huge amount of information stored in such event logs. The main objective not only consists of finding out if drivers workplan complies with the HoS regulation, but also summarising their activities in a concise but representative way. Additionally, these underlying patterns in the event log could be analysed in order to discover driving styles which could make possible the suggestion of routes or tasks more aligned to the driver preferences.

The creation of driver profiles based on driving styles with HoS can be extremely useful for managers, as they could assign transport routes to the most appropriate drivers, given the length of the route and the proximity of the deadline. For example, drivers who maximise their driving hours could be preferred for long distance routes and drivers who tend to take split rest to on-city deliveries.

Therefore, in this paper we present a method that, starting from real event logs extracted from a tachograph device, 1) labels driver activities according to the HoS regulation, 2) identifies infractions and their cause, 3) extract summarised information about the log while clustering driving sequences based on similar behaviour patterns, and 4) group drivers by similarity of those clustered patterns. As a results, experts are provided with an understandable analysis of what the driver has been doing in multiple levels of granularity, from a detailed description of the activities and infractions under the HoS regulation to a categorisation with similar tendencies.

The remainder of this paper shows, firstly, a description of the problem addressed and some background concepts to it. Then, we present the methodology of the approach, followed by details of experimentation conducted over a proof of concept of the application. Finally, we conclude discussing related and future work.


\section{Problem Description}

\begin{figure}[t]
    \centering
    \includegraphics[width=.7\columnwidth]{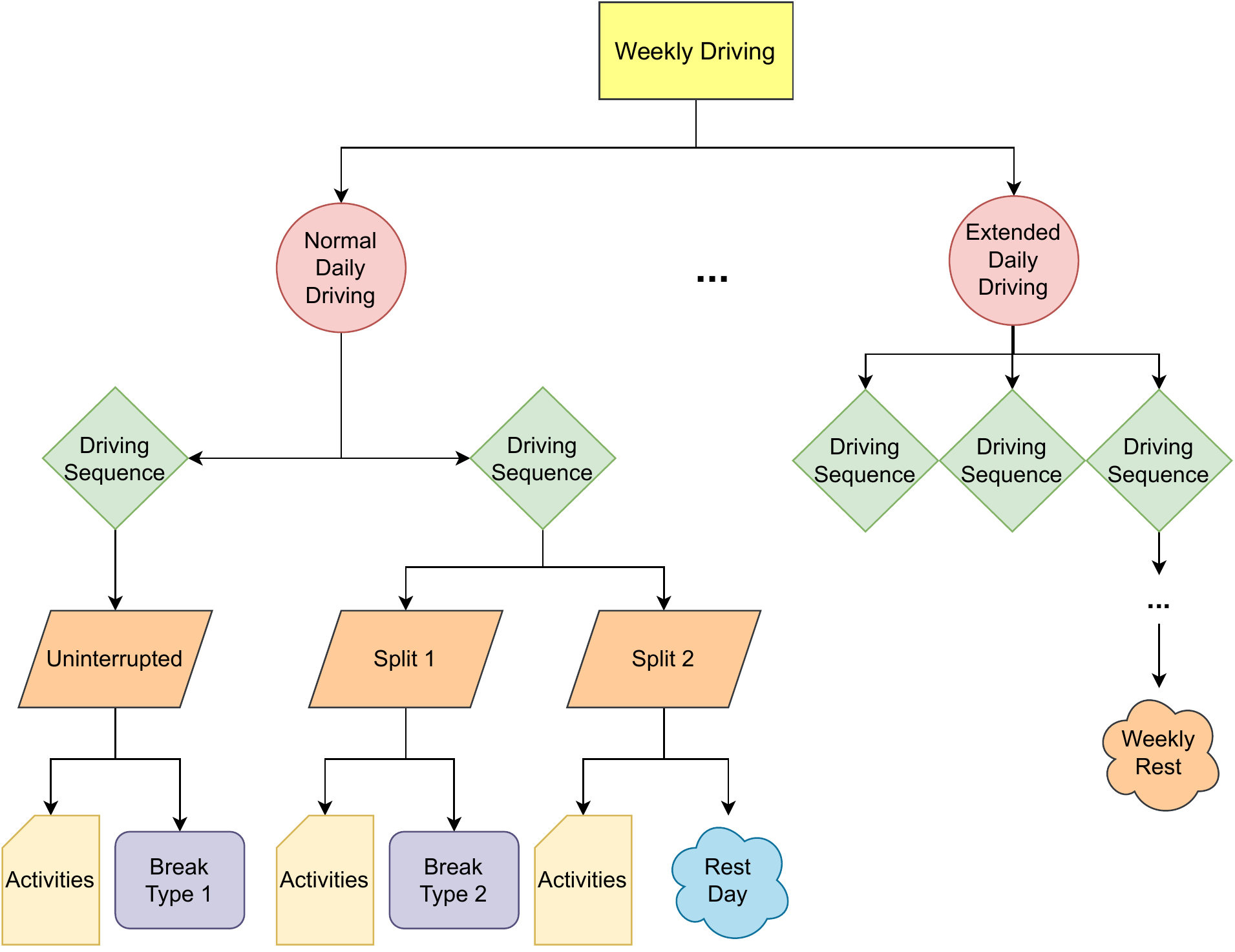}
    \caption{Partial example of the HoS tree. At the upmost level, a weekly driving period is formed by several daily periods, which must end with a weekly rest. Similarly, daily driving periods are separated by daily rests, and according to the accumulative hours of driving in them can be classified as Normal Daily Driving periods (up to 9 hours) or Extended Daily Driving periods (more than 9 hours). Because each driving sequence should not surpass 4.5 hours of driving time, they can be distinguished by the number of driving sequences in them.}
    \label{fig:hos}
\end{figure}

We are collaborating with a company which provides decision support based on
prediction services to its customers. Ultimately they want to help them govern the behaviour of their drivers by predicting whether a driver is close to committing an infraction, as well as characterising drivers according to their driving style with respect to the HoS regulation.

They handed us tachograph logs of multiple drivers with thousands of activities and asked us to develop a system to analyse driver behaviour. Due to the regulation imposing additional difficulties at interpreting the data, and the high volume that is constantly being generated, experts cannot interpret directly the original tachograph logs and require summarisation of what a driver has been doing during that period of time to make business decisions. A tachograph \citep{baldini2018regulated} is an automated recording device fitted into a vehicle that extracts information from the driving activities such as speed, duration and distance.

Our dataset represented an event log where every activity is a tuple $(id,\;start,\;end,\;dur,\;a)$, each component referring to: driver identifier $id$; $start$ and $end$ timestamps; activity duration $dur$; and activity identifier $a$, respectively. A value for $a$ is any of the labels $[Driving,Other,Break,Idle]$ meaning that the driver is either Driving, performing Another Work, at Break or Idle during $dur$ minutes, between $start$ and $end$. The semantics of each event is completed with the definitions provided by the HoS regulation, which are detailed in the following paragraphs.

Although the HoS standard is applied in several countries, in this work we focus on the European Union regulation (EC) No 561/2006, which has been extensively analysed in \citep{goel2013hours,meyer_european_2011}. The basic terms refer to four types of driver activities as \textit{break} (short period for recuperation), \textit{rest} (free disposal period with enough time to sleep), \textit{driving} (time during which the driver is operating a vehicle) and \textit{other work} (time devoted to any work except driving, like loading).

These activities are hierarchically grouped up to weekly intervals, based on the duration of the events contained in them. To ease the explanation of this article we are referring at the whole structures as \textbf{HoS trees}. In Figure \ref{fig:hos} we exemplify a portion of a HoS tree displaying a \textbf{N}ormal \textbf{D}aily \textbf{D}riving (NDD) period on the first day and a \textbf{E}xtended \textbf{D}aily \textbf{D}riving (EDD) period on the last.

At the lower levels activities are joined in different types of \textit{driving sequences}. A basic driving sequence is composed of a totally ordered set of the elements of $[Driving,Other,Break,Idle]$ constrained so that the duration of any $Break$ is less than 15 minutes. More constraints are defined over the duration of the rests and breaks, and over the accumulated duration of driving sequences.

The regulation provides a set of basic and optional rules, should the former not be satisfied, thus allowing more flexibility to generate and interpret driving schedules under such constraints. For example, either a break of 45 min has to be taken after 4.5 hours of accumulated driving or it can be taken split in two parts of at least 15 min and 30 min respectively. This feature is good for drivers since it provides flexibility to their work, but complicates the interpretability of what they are doing. The regulation also defines additional constraints (for example, the maximum number of occurrences of a reduced rest in a weekly driving period), and relationships between the different types of sub-sequences, as well as their internal structure.

\begin{figure*}[!t]
	\centering
    \includegraphics[width=\columnwidth]{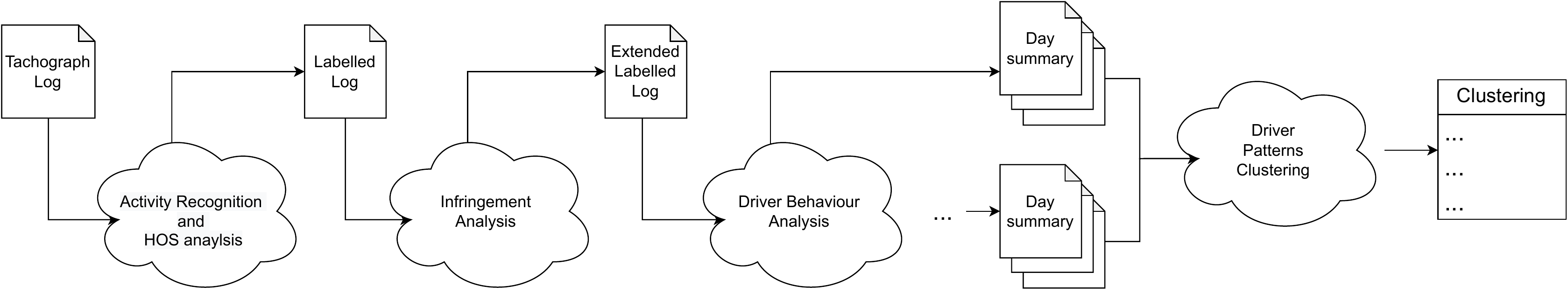}
    \caption{General overview of our approach.}
    \label{fig:overview}
\end{figure*}

\section{Background}

Automated planning \citep{automatedPlanning} is a branch of A.I. concerned with the study of agent acting techniques. However, its uses can be broaden and as we show in this paper planning can also be applied to recognition tasks.

Two elements are required in a planning environment: (i) the action models existing in the world, referred as \textbf{domain}; and (ii) a description of the initial state of the world, the objects involved in it and the desired goals, called \textbf{problem}. These two inputs are provided to a \textbf{planner}, a search-based algorithm that determines the plan (sequence of actions) that achieve the goals from the starting state.

Our proposed methodology employs hierarchical planning, more commonly referred as Hierarchical Task Networks (HTN). HTNs forms a branch of classical planning where the domain can be decomposed in hierarchical structures of tasks/subtasks, with low level tasks representing temporally annotated actions, and compound tasks representing temporal ordering strategies between those actions.


\section{Application Overall Description}

To solve the problem of explaining and summarising a driver's tachograph log and its compliance with the HoS regulation we propose a modular architecture divided in three main components, as seen in Figure \ref{fig:overview}:

\begin{itemize}
    \item First, an initial planning process to label the input tachograph log according to the HoS regulation.
    \item Then, a system to identify and explain the causes of driver infractions extending the previous labelled log.
    \item Thirdly, a module to analyse driver behaviour via summarisation of driving sequences.
    \item Lastly, summarised driving days are used as training data to clusterize drivers by similar driving patterns.
\end{itemize}

The following subsections provide a detailed explanation of each component.

\subsection{Labelling Driver Activities}\label{sec:labelling}

\begin{figure}[!t]
	\centering
    \includegraphics[width=.8\columnwidth]{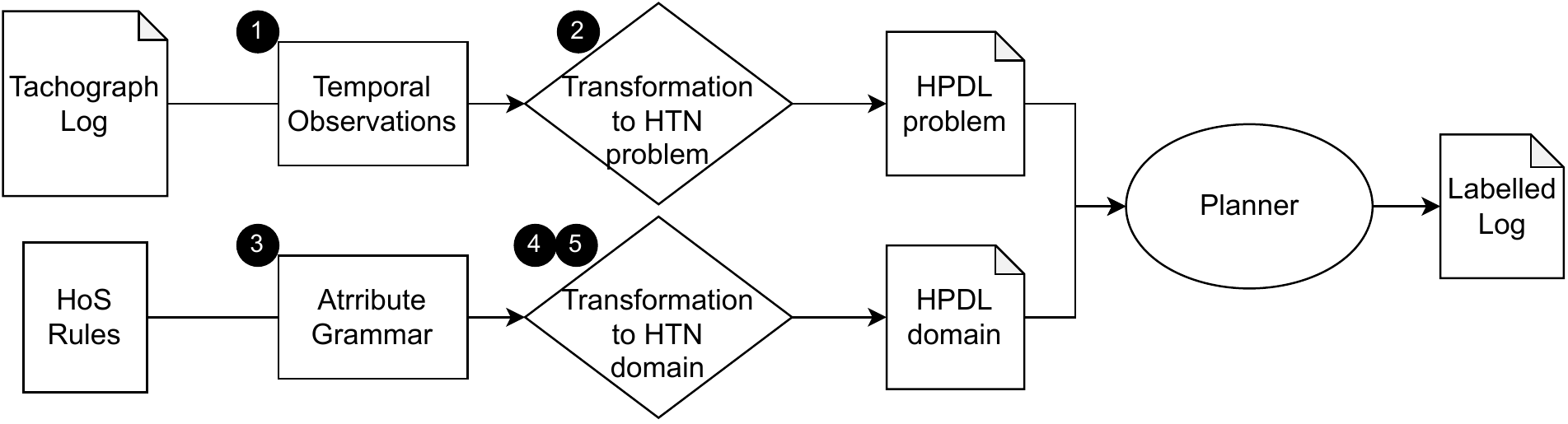}
    \caption{Labelling process for a tachograph log.}
    \label{fig:labelling}
\end{figure}

\begin{table*}[t]
    \caption{Labelling output for legal activities. This example shows the second (and last) driving sequence in a normal daily driving period, where the required break has been taken in two parts, a small break in the first split and a second one extended as a daily rest.}
    \label{tab:legal}
\renewcommand{\arraystretch}{1.2}
\resizebox{\textwidth}{!}{%
\begin{tabular}{|lllll|lllllll|}
\hline
\multicolumn{5}{|c|}{Original Log}                                                            & \multicolumn{7}{c|}{Annotated Labels}                                                                                      \\ \hline
\textbf{Driver} & \textbf{Start}   & \textbf{End}     & \textbf{Duration} & \textbf{Activity} & \textbf{Week} & \textbf{Day} & \textbf{DayType} & \textbf{Sequence} & \textbf{BreakType} & \textbf{Token} & \textbf{Legal} \\ \hline
driver1                          & 11/01/2017 17:33                & 11/01/2017 17:37              & 4                                  & Driving                            & \multicolumn{1}{c|}{\multirow{9}{*}{1}}            & \multicolumn{1}{c|}{\multirow{9}{*}{4}}           & \multicolumn{1}{c|}{\multirow{9}{*}{ndd}}             & \multicolumn{1}{c|}{\multirow{9}{*}{second}}           & \multicolumn{1}{c|}{\multirow{2}{*}{split\_1}}          & \multicolumn{1}{l|}{A}          & yes                             \\ \cline{1-5} \cline{11-12} 
driver1                          & 11/01/2017 17:37                & 11/01/2017 18:16              & 39                                 & Break                              & \multicolumn{1}{c|}{}                              & \multicolumn{1}{c|}{}                             & \multicolumn{1}{c|}{}                                 & \multicolumn{1}{c|}{}                                  & \multicolumn{1}{c|}{}                                   & \multicolumn{1}{l|}{B\_T2}      & yes                             \\ \cline{1-5} \cline{10-12} 
driver1                          & 11/01/2017 18:16                & 11/01/2017 18:17              & 1                                  & Driving                            & \multicolumn{1}{c|}{}                              & \multicolumn{1}{c|}{}                             & \multicolumn{1}{c|}{}                                 & \multicolumn{1}{c|}{}                                  & \multicolumn{1}{c|}{\multirow{7}{*}{split\_2}}          & \multicolumn{1}{l|}{A}          & yes                             \\ \cline{1-5} \cline{11-12} 
driver1                          & 11/01/2017 18:17                & 11/01/2017 18:25              & 8                                  & Other                              & \multicolumn{1}{c|}{}                              & \multicolumn{1}{c|}{}                             & \multicolumn{1}{c|}{}                                 & \multicolumn{1}{c|}{}                                  & \multicolumn{1}{c|}{}                                   & \multicolumn{1}{l|}{A}          & yes                             \\ \cline{1-5} \cline{11-12} 
driver1                          & 11/01/2017 18:25                & 11/01/2017 19:54              & 89                                 & Driving                            & \multicolumn{1}{c|}{}                              & \multicolumn{1}{c|}{}                             & \multicolumn{1}{c|}{}                                 & \multicolumn{1}{c|}{}                                  & \multicolumn{1}{c|}{}                                   & \multicolumn{1}{l|}{A}          & yes                             \\ \cline{1-5} \cline{11-12} 
driver1                          & 11/01/2017 19:54                & 11/01/2017 19:57              & 3                                  & Break                              & \multicolumn{1}{c|}{}                              & \multicolumn{1}{c|}{}                             & \multicolumn{1}{c|}{}                                 & \multicolumn{1}{c|}{}                                  & \multicolumn{1}{c|}{}                                   & \multicolumn{1}{l|}{B\_T0}      & yes                             \\ \cline{1-5} \cline{11-12} 
driver1                          & 11/01/2017 19:57                & 11/01/2017 19:58              & 1                                  & Driving                            & \multicolumn{1}{c|}{}                              & \multicolumn{1}{c|}{}                             & \multicolumn{1}{c|}{}                                 & \multicolumn{1}{c|}{}                                  & \multicolumn{1}{c|}{}                                   & \multicolumn{1}{l|}{A}          & yes                             \\ \cline{1-5} \cline{11-12} 
driver1                          & 11/01/2017 19:58                & 11/01/2017 20:01              & 3                                  & Other                              & \multicolumn{1}{c|}{}                              & \multicolumn{1}{c|}{}                             & \multicolumn{1}{c|}{}                                 & \multicolumn{1}{c|}{}                                  & \multicolumn{1}{c|}{}                                   & \multicolumn{1}{l|}{A}          & yes                             \\ \cline{1-5} \cline{11-12} 
driver1                          & 11/01/2017 20:01                & 12/01/2017 07:06              & 665                                & Break                              & \multicolumn{1}{c|}{}                              & \multicolumn{1}{c|}{}                             & \multicolumn{1}{c|}{}                                 & \multicolumn{1}{c|}{}                                  & \multicolumn{1}{c|}{}                                   & \multicolumn{1}{l|}{DR\_T1}     & yes                             \\ \hline
\end{tabular}%
}
\end{table*}

To label our logs with HoS terms we employ our previously developed methodology proposed in \citep{vehits22}, where a HTN domain serves to both recognise and tag activities from a tachograph log. We provide a brief summary below, but we refer the reader to the original paper for an in depth explanation of the methodology. The overall steps of this system, represented in Figure \ref{fig:labelling}, are:

\begin{enumerate}
    \item Generate a set of ordered temporal observations from the tachograph activity log, which are part of the initial state of a HTN problem.
    \item Represent the recognition of a driver activity as a temporal HTN problem, where an activity is added to the plan if (i) the temporal information of the activity is consistent with the domain, and (ii) the temporal constraints of the activity are consistent with the rest of temporal constraints of the actions already added to the plan.
    \item Codify a HoS tree in an attribute grammar \citep{knuth1968semantics} as an intermediate representation, with HoS rules as productions.
    \item Translate the grammar into a temporal HTN domain, aimed at representing the parsing of the activity log as a HTN problem where (i) terminal symbols are recognised as temporal events and (ii) nonterminal symbols are recognised according to grammar rules.
    \item Extend the domain to both recognise and label activities from the log to be easily interpretable. The resulting log contains five new labels according to the contexts \textit{DayType} (Normal or Extended Daily Driving period), \textit{Sequence} (if the activity belongs to the first, second or third sequence in the day), \textit{BreakType} (if breaks are taken in one or two parts), \textit{Token} (the type of activity at the lowest level of in the HoS tree\footnote{Many types of categories exist at the lowest level, based on the duration of the action. For example, \textit{A} indicates a working activity, \textit{B\_T0} a break of less than 15 minutes, \textit{DR\_T1} a daily rest of more than 11 hours, and \textit{WR\_T1} a weekly rest with more than 45 hours.}) and \textit{Legal} (whether the activity complies or not with the regulation), as well as two counter columns for the day and the week processed. An output example can be seen in Table \ref{tab:legal}.
\end{enumerate}

In summary, the recognition problem is solved with a planning process where the domain walks through an activity log and its internal HTN structure simultaneously, the latter codifying the HoS tree. If activities comply with the temporal and formal restrictions they are labelled with the appropriate terms, in other case contexts are tagged as unrecognised.

Nevertheless, the domain is designed to label as many contexts as possible. If a higher (more general) context cannot be identified, the domain still attempts to identify lower contexts before ignoring the action. That means that when a bigger sequence cannot be grouped and labelled together (\textit{e.g.} when the driver exceeds the maximum number of driving hours and the \textit{DayType} column cannot be tagged), the domain tries to tag smaller sequences with their corresponding label. An example is shown below in Table \ref{tab:illegal}, where although DayType and Sequence tags could not be recognised the system still identifies both BreakType splits and includes the appropriate labels, as well as the correct Token contexts.

\subsection{Explaining Infringements}

\begin{figure}[!t]
	\centering
    \includegraphics[width=.72\columnwidth]{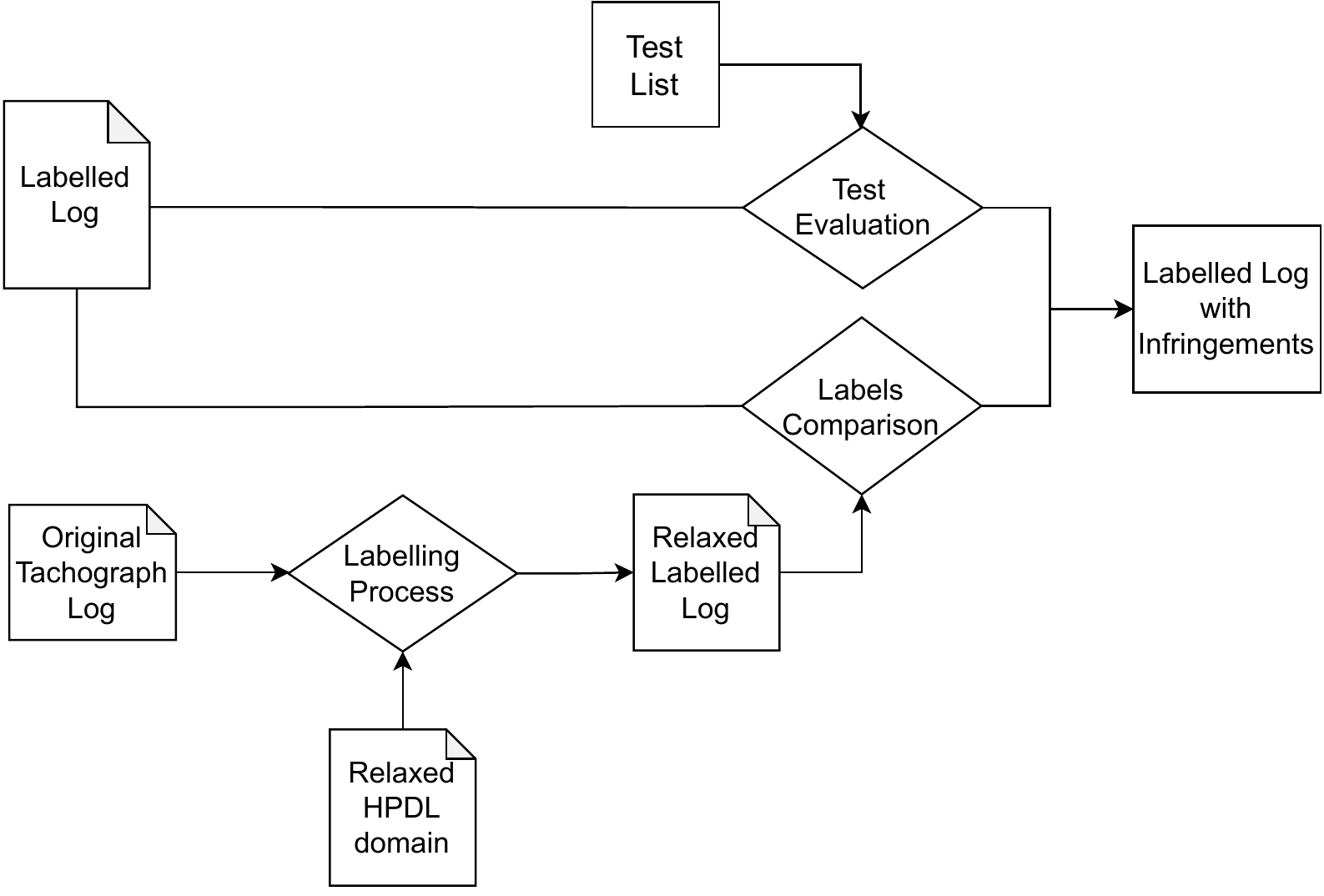}
    \caption{Infringement analysis process for a labelled log.}
    \label{fig:infringements}
\end{figure}

The previous recognition process labels the tachograph log considering the terms defined by the HoS regulation, its compliance with it and details of their position in the HoS tree. However, when drivers commit infractions this system by itself cannot provide an explanation of the cause and the exact root activity, due to the fact that planning techniques rely on backtracking (that is, the ability to retract while exploring the planning graph) and there is not a simple way to distinguish between a genuine backtracking step while walking through the HTN domain or a forced one by an illegal activity in the log.

Therefore we found a need to further analyse the labelled log and explain these information to users without requiring them to inspect all activities not recognised in the log. We solved this problem from two perspectives, each one concerned with different kinds of violations, which are explained in the following subsections. Figure \ref{fig:infringements} shows an overview of the approaches.

\begin{table}[t]
\caption{Tests applied to driving sequences in the log in order to identify infringement causes.\label{tab:restrictions}}
\centering
\begin{tabular}{c|c}
\textbf{Test} & \textbf{Infraction type} \\
\hline
dt\_seq $>$ 4.5h & Excessive Driving without breaks \\
\hline
\thead{dt\_day $>$ 9h \\ $\land$ \\ EDDs\_this\_week $>$ 2} & Excessive Driving in day (NDD) \\
\hline
dt\_day $>$ 10h & Excessive Driving in day (EDD) \\
\hline
\thead{Token day before = DR\_T3 \\ $\land$ \\ Token = $\neg$ (DR\_T4 or WR)} & Missing other half of split daily rest \\
\hline
\thead{Token = DR or WR \\ $\land$ \\ Legal = No \\ $\land$ \\ \thead{Remaining \\ contexts} = $\neg$ Unknown} & Rest past the daily/weekly deadline \\
\end{tabular}
\end{table}

\begin{table*}[t]
\caption{Labelling output example for illegal activities and the infraction detected by the tests list.}
    \label{tab:illegal}
\renewcommand{\arraystretch}{1.2}
\resizebox{\textwidth}{!}{%
\begin{tabular}{|lllll|lllllll|c|}
\hline
\multicolumn{5}{|c|}{Original Log}                                                            & \multicolumn{8}{c|}{Annotated Labels}                                                                                                                                     \\ \hline
\textbf{Driver} & \textbf{Start}   & \textbf{End}     & \textbf{Duration} & \textbf{Activity} & \textbf{Week} & \textbf{Day} & \textbf{DayType} & \textbf{Sequence} & \textbf{BreakType} & \textbf{Token} & \textbf{Legal} & \multicolumn{1}{c|}{\textbf{Infraction}}     \\ \hline
driver39 & 10/01/2017 12:12 & 10/01/2017 14:17 & 125 & Driving & \multicolumn{1}{c|}{\multirow{12}{*}{1}} & \multicolumn{1}{c|}{\multirow{12}{*}{5}} & \multicolumn{1}{c|}{\multirow{4}{*}{unkown}} & \multicolumn{1}{c|}{\multirow{4}{*}{unkown}} & \multicolumn{1}{c|}{\multirow{2}{*}{split\_1}}      & \multicolumn{1}{c|}{A}      & \multicolumn{1}{c|}{no}  & \multirow{12}{*}{Surpassed NDD driving time} \\ \cline{1-5} \cline{11-12}
driver39 & 10/01/2017 14:17 & 10/01/2017 14:40 & 23  & Break   & \multicolumn{1}{c|}{}                    & \multicolumn{1}{c|}{}                    & \multicolumn{1}{c|}{}                        & \multicolumn{1}{c|}{}                        & \multicolumn{1}{c|}{}                               & \multicolumn{1}{c|}{B\_T2}  & \multicolumn{1}{c|}{no}  &                                              \\ \cline{1-5} \cline{10-12}
driver39 & 10/01/2017 14:40 & 10/01/2017 16:52 & 132 & Driving & \multicolumn{1}{c|}{}                    & \multicolumn{1}{c|}{}                    & \multicolumn{1}{c|}{}                        & \multicolumn{1}{c|}{}                        & \multicolumn{1}{c|}{\multirow{2}{*}{split\_2}}      & \multicolumn{1}{c|}{A}      & \multicolumn{1}{c|}{no}  &                                              \\ \cline{1-5} \cline{11-12}
driver39 & 10/01/2017 16:52 & 10/01/2017 17:25 & 33  & Break   & \multicolumn{1}{c|}{}                    & \multicolumn{1}{c|}{}                    & \multicolumn{1}{c|}{}                        & \multicolumn{1}{c|}{}                        & \multicolumn{1}{c|}{}                               & \multicolumn{1}{c|}{B\_T3}  & \multicolumn{1}{c|}{no}  &                                              \\ \cline{1-5} \cline{8-12}
driver39 & 10/01/2017 17:25 & 10/01/2017 20:27 & 182 & Driving & \multicolumn{1}{c|}{}                    & \multicolumn{1}{c|}{}                    & \multicolumn{1}{c|}{\multirow{8}{*}{ndd}}    & \multicolumn{1}{c|}{\multirow{6}{*}{first}}  & \multicolumn{1}{c|}{\multirow{2}{*}{split\_1}}      & \multicolumn{1}{c|}{A}      & \multicolumn{1}{c|}{yes} &                                              \\ \cline{1-5} \cline{11-12}
driver39 & 10/01/2017 20:27 & 10/01/2017 20:42 & 15  & Break   & \multicolumn{1}{c|}{}                    & \multicolumn{1}{c|}{}                    & \multicolumn{1}{c|}{}                        & \multicolumn{1}{c|}{}                        & \multicolumn{1}{c|}{}                               & \multicolumn{1}{c|}{B\_T2}  & \multicolumn{1}{c|}{yes} &                                              \\ \cline{1-5} \cline{10-12}
driver39 & 10/01/2017 20:42 & 10/01/2017 21:54 & 72  & Driving & \multicolumn{1}{c|}{}                    & \multicolumn{1}{c|}{}                    & \multicolumn{1}{c|}{}                        & \multicolumn{1}{c|}{}                        & \multicolumn{1}{c|}{\multirow{4}{*}{split\_2}}      & \multicolumn{1}{c|}{A}      & \multicolumn{1}{c|}{yes} &                                              \\ \cline{1-5} \cline{11-12}
driver39 & 10/01/2017 21:54 & 10/01/2017 21:59 & 5   & Break   & \multicolumn{1}{c|}{}                    & \multicolumn{1}{c|}{}                    & \multicolumn{1}{c|}{}                        & \multicolumn{1}{c|}{}                        & \multicolumn{1}{c|}{}                               & \multicolumn{1}{c|}{B\_T0}  & \multicolumn{1}{c|}{yes} &                                              \\ \cline{1-5} \cline{11-12}
driver39 & 10/01/2017 21:59 & 10/01/2017 22:00 & 1   & Driving & \multicolumn{1}{c|}{}                    & \multicolumn{1}{c|}{}                    & \multicolumn{1}{c|}{}                        & \multicolumn{1}{c|}{}                        & \multicolumn{1}{c|}{}                               & \multicolumn{1}{c|}{A}      & \multicolumn{1}{c|}{yes} &                                              \\ \cline{1-5} \cline{11-12}
driver39 & 10/01/2017 22:00 & 10/01/2017 22:37 & 37  & Break   & \multicolumn{1}{c|}{}                    & \multicolumn{1}{c|}{}                    & \multicolumn{1}{c|}{}                        & \multicolumn{1}{c|}{}                        & \multicolumn{1}{c|}{}                               & \multicolumn{1}{c|}{B\_T3}  & \multicolumn{1}{c|}{yes} &                                              \\ \cline{1-5} \cline{9-12}
driver39 & 10/01/2017 22:37 & 10/01/2017 23:21 & 44  & Driving & \multicolumn{1}{c|}{}                    & \multicolumn{1}{c|}{}                    & \multicolumn{1}{c|}{}                        & \multicolumn{1}{c|}{\multirow{2}{*}{second}} & \multicolumn{1}{c|}{\multirow{2}{*}{uninterrupted}} & \multicolumn{1}{c|}{A}      & \multicolumn{1}{c|}{yes} &                                              \\ \cline{1-5} \cline{11-12}
driver39 & 10/01/2017 23:21 & 11/01/2017 08:53 & 572 & Break   & \multicolumn{1}{c|}{}                    & \multicolumn{1}{c|}{}                    & \multicolumn{1}{c|}{}                        & \multicolumn{1}{c|}{}                        & \multicolumn{1}{c|}{}                               & \multicolumn{1}{c|}{DR\_T2} & \multicolumn{1}{c|}{yes} &                                              \\ \hline
\end{tabular}%
}
\end{table*}

\subsubsection{Test evaluation}
On one hand we represent rules from the HoS regulation as tests and applied them to the sequences the labelling process found unrecognisable events (i.e., those missing at least a label). These tests, as exemplified in the left part of Table \ref{tab:restrictions}, codify limits and restrictions in the duration of driving sequences and breaks. Whenever a test flags a sequence the system marks it and provides an explanation of the infringement, as seen in Table \ref{tab:illegal}.

Tests takes the form of logic constraints 
\begin{align}
    f(a_{start}, a_{end}) \ o \ V
\end{align}
being:

\begin{itemize}
    \item $f$ a function applied over the sequence defined between activities $a_{start}$ and $a_{end}$ (\textit{e.g.} sum, context value).
    \item $o$ a logic operator. 
    \item $V$ either the value of a context (\textit{e.g.} Token, DayType), a scalar or a duration.
\end{itemize}

As an example, the first constraint in Table \ref{tab:restrictions} could be rewritten as $duration({seq}_{start}, {seq}_{end}) > 4.5h$.

It is important to note that to correctly identify the infraction some tests may consider not only the illegal activities but also prior activities of other days, a situation frequently present in reduced breaks and rests, where sometimes compensation breaks are not fulfilled. Therefore the interval of activities checked by the tests depends on the test itself.

Because tests are encoded as logic constraints, it is easy to extend the system with additional expert provided rules or modify them if the regulation changes.

\subsubsection{Re-labelling}
A second approach consists of re-labelling the log using a domain with relaxed duration intervals, that is, the limits imposed by the regulation are softened (\textit{e.g.} maximum driving time or minimum break time are enlarged up and down) and the system looks for changes between the new log and the original tagged log.

This process helps to discover infringements caused by a slightly borderline duration, like the driver surpassing (probably unconsciously) the restriction by a small amount. These kind of situations are not easily identified by the tests, due to the fact that the activity by itself could still be legal but labelled differently, becoming an infraction later on.

For example, a driver could surpass the maximum limit for a pause before being considered a break by a few minutes without noticing, and proceeding like a break has not been consumed. As a consequence, that action will be valid, but after the next breaks infractions may arise because the driver is not following its plan as expected, and such actions may not fit correctly under the HoS tree.

If the violation is related with this type of mistake, the new relabelled log will contain less illegal sequences than the original and we can compare the \textit{Token} contexts (concerning the type of activity at the lowest level in the HoS tree) to understand which changes make the sequence legal. Table \ref{tab:relaxed} shows an example with a driver exceeding the break time by two minutes.

Therefore, this method allows us to (a) discover new infringements not considered by the test list, and (b) analyse how the activity should have been to avoid infractions.

\begin{table}[t]
    \caption{Identifying infringements with a relaxed domain. In this example the fourth activity surpasses by one minute the duration limit to be considered B\_T0, making the whole sequence illegal.}
    \label{tab:relaxed}
    \centering
\renewcommand{\arraystretch}{1.2}
\resizebox{0.7\textwidth}{!}{%
\begin{tabular}{|l|l|l|l|l|l|l|}
\hline
\multicolumn{7}{|c|}{Original Labelled Log}                                                                                         \\ \hline
\textbf{Duration} & \textbf{Activity} & \textbf{DayType} & \textbf{Sequence} & \textbf{BreakType} & \textbf{Token} & \textbf{Legal} \\ \hline
57                                 & \multicolumn{1}{l|}{Driving}       & \multicolumn{1}{c|}{\multirow{4}{*}{unknown}}         & \multicolumn{1}{c|}{\multirow{4}{*}{unknown}}          & \multicolumn{1}{c|}{\multirow{4}{*}{split\_1}}          & A                               & no                              \\ \cline{1-2} \cline{6-7} 
3                                  & \multicolumn{1}{l|}{Break}         & \multicolumn{1}{c|}{}                                 & \multicolumn{1}{c|}{}                                  & \multicolumn{1}{c|}{}                                   & B\_T0                           & no                              \\ \cline{1-2} \cline{6-7} 
2                                  & \multicolumn{1}{l|}{Driving}       & \multicolumn{1}{c|}{}                                 & \multicolumn{1}{c|}{}                                  & \multicolumn{1}{c|}{}                                   & A                               & no                              \\ \cline{1-2} \cline{6-7} 
16                                 & \multicolumn{1}{l|}{Break}         & \multicolumn{1}{c|}{}                                 & \multicolumn{1}{c|}{}                                  & \multicolumn{1}{c|}{}                                   & \textbf{B\_T2} & no                              \\ \hline \hline
\multicolumn{7}{|c|}{Relaxed Labelled Log}                                                                                          \\ \hline
\textbf{Duration} & \textbf{Activity} & \textbf{DayType} & \textbf{Sequence} & \textbf{BreakType} & \textbf{Token} & \textbf{Legal} \\ \hline
57                                 & \multicolumn{1}{l|}{Driving}       & \multicolumn{1}{c|}{\multirow{4}{*}{ndd}}             & \multicolumn{1}{c|}{\multirow{4}{*}{first}}            & \multicolumn{1}{c|}{\multirow{4}{*}{split\_1}}          & A                               & yes                             \\ \cline{1-2} \cline{6-7} 
3                                  & \multicolumn{1}{l|}{Break}         & \multicolumn{1}{c|}{}                                 & \multicolumn{1}{c|}{}                                  & \multicolumn{1}{c|}{}                                   & B\_T0                           & yes                             \\ \cline{1-2} \cline{6-7} 
2                                  & \multicolumn{1}{l|}{Driving}       & \multicolumn{1}{c|}{}                                 & \multicolumn{1}{c|}{}                                  & \multicolumn{1}{c|}{}                                   & A                               & yes                             \\ \cline{1-2} \cline{6-7} 
16                                 & \multicolumn{1}{l|}{Break}         & \multicolumn{1}{c|}{}                                 & \multicolumn{1}{c|}{}                                  & \multicolumn{1}{c|}{}                                   & \textbf{B\_T0} & yes                             \\ \hline
\end{tabular}%
}
\end{table}


\subsection{Analysing Driver Behaviour}\label{sec:clustering}

The two previous steps provide a way to understand a driver log and its compliance with the HoS regulation. However, experts are usually responsible of dozens of drivers and its not feasible to analyse the substantial logs of each one of them in order to detect problematic tendencies.

Therefore, we developed a module that clusters behaviour patterns in driver activities and summarises each cluster with expert knowledge. This method helps to separate standard driving days from unusual ones without inspecting the driver log, and let users concentrate their efforts in analysing only the problematic sequences.

In order to do that, we considered our problem as an NLP (Natural Language Processing) task, where activities from the log are treated as words and daily sequences as documents. That way we can employ NLP oriented techniques to transform sequences of varying length into fixed dimensions and measure similarity between them.

\begin{figure}[!t]
	\centering
    \includegraphics[width=.8\columnwidth]{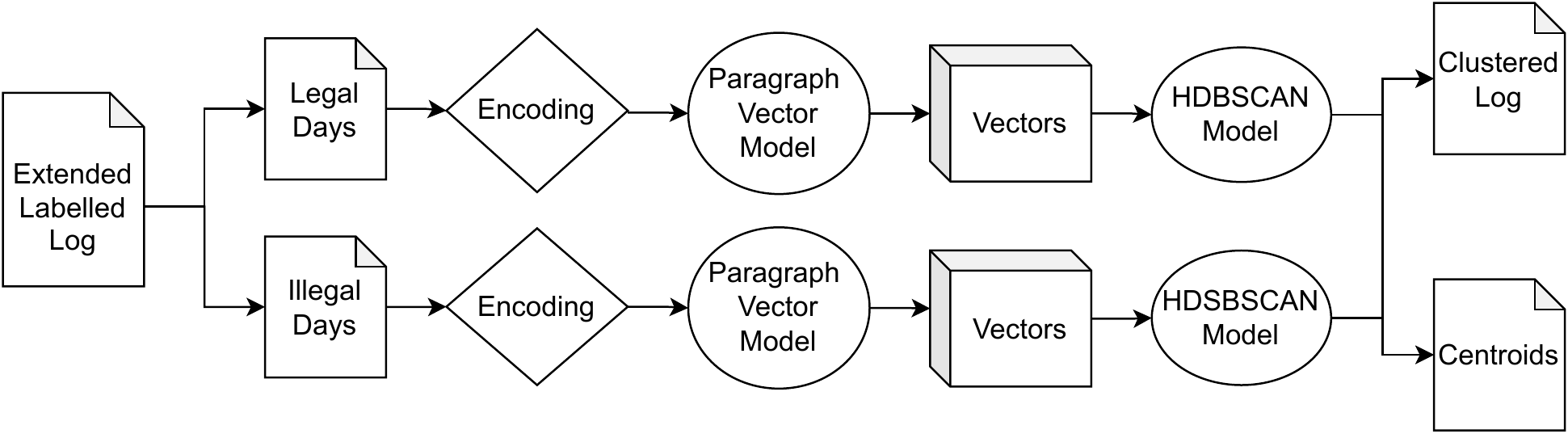}
    \caption{Clustering process for a labelled log.}
    \label{fig:nlp}
\end{figure}

Figure \ref{fig:nlp} shows an overview of the process, consisting of the following steps:

\begin{enumerate}
    \item First, a preprocessing step is applied in which the dataset is split in two parts depending on whether the days contains or not illegal activities. The reason behind this process is that an infraction recognition process is already provided by the previous module, and thus there is no need for our clustering model to learn to distinguish between legal and illegal sequences. On the contrary, we are providing a prior separation to help the model extract more interesting patterns that are not related with the legality of the sequence.

    \item The subset of labelled columns (\textit{i.e.} contexts) that describe the action from an overall point of view are selected, these are \textit{(Activity, DayType, BreakType, Token)}. For the illegal subset, the \textit{Infraction} column is also included to generate clusters and centroids associated with already identified infringement. Specific details about duration and timestamps are not relevant to summarise the days. Nevertheless, some of the information they provided is encoded in the labels, as it is used by the labelling process. This step could be consider as cleaning a document prior an NLP topic categorisation task.
    
    \item Because columns contains categorical features not suitable for computation they are transformed into numerical, and then joined together using a special character as a separator. After this step we can consider each entry in our log as a word.
    
    \item Both previous steps are repeated for each activity in our dataset, and activities of the same day are grouped into documents. As a result, we have a collection of documents each one encoding the activities in a driving day sequence as words. 
    
    \item We then use Paragraph Vector \citep{doc2vec} (also known as Doc2Vec) models to obtain dense representations of fixed dimensions\footnote{Other techniques like Word2Vec or Bag of Words could be used, but we considered the paragraph weight extracted by Paragraph Vector a useful source of information in our task.}. Although one model could be trained for both data splits (and reasonable so, as both encodings are subset of the same language), we obtained better results finetuning one model for each split, but ultimately both transforming a document into a 200 sized output vector.

    \item  The resulting representations are now suitable for clustering techniques. We obtained our best results using HDBSCAN \citep{hdbscan} thanks to its robustness to noise, and choosing the number of clusters based on both expert knowledge and metrics results (Silhouette Coefficient, Calinski-Harabasz and Davies-Bouldin indexs), setting on a final value of 8 clusters for legal data and 7 for days with infractions. In the next section we display a comparative analysis of other techniques under this data.
    
    \item Lastly, days are clustered and presented with the decoded centroids, which are described by an expert with a meaningful description, as shown in Table \ref{tab:clustered}.
\end{enumerate}

\begin{table}[t]
    \caption{Partial output of the clustering process. The system identifies the most similar centroid to the input sequence and the description associated with it.}
    \label{tab:clustered}
    \centering
\renewcommand{\arraystretch}{1.2}
\resizebox{0.7\textwidth}{!}{%
\begin{tabular}{|llllllc|}
\hline
\multicolumn{7}{|c|}{Labelled Log}                                                                                                                                                                                                                          \\ \hline
\multicolumn{1}{|c}{\textbf{Activity}} & \multicolumn{1}{c}{\textbf{DayType}} & \multicolumn{1}{c}{\textbf{Sequence}} & \multicolumn{1}{c}{\textbf{BreakType}} & \multicolumn{1}{c}{\textbf{Token}} & \multicolumn{1}{c}{\textbf{Legal}} & \textbf{Cluster} \\ \hline
\multicolumn{1}{|l|}{Driving}  & \multicolumn{1}{c|}{\multirow{3}{*}{ndd}} & \multicolumn{1}{c|}{\multirow{3}{*}{unique}} & \multicolumn{1}{c|}{\multirow{3}{*}{uninterrupted}} & \multicolumn{1}{l|}{A}      & \multicolumn{1}{l|}{yes}   & \multirow{3}{*}{2}                \\ \cline{1-1} \cline{5-6}
\multicolumn{1}{|l|}{Other}    & \multicolumn{1}{c|}{}                     & \multicolumn{1}{c|}{}                        & \multicolumn{1}{c|}{}                               & \multicolumn{1}{l|}{A}      & \multicolumn{1}{l|}{yes}   &                                   \\ \cline{1-1} \cline{5-6}
\multicolumn{1}{|l|}{Break}    & \multicolumn{1}{c|}{}                     & \multicolumn{1}{c|}{}                        & \multicolumn{1}{c|}{}                               & \multicolumn{1}{l|}{DR\_T1} & \multicolumn{1}{l|}{yes}   &                                   \\ \hline
\hline
\multicolumn{7}{|c|}{Most similar centroid}                                                                                                                                                                                                                 \\ \hline
\multicolumn{1}{|c}{\textbf{Activity}} & \multicolumn{1}{c}{\textbf{DayType}} & \multicolumn{1}{c}{\textbf{Sequence}} & \multicolumn{1}{c}{\textbf{BreakType}} & \multicolumn{1}{c}{\textbf{Token}} & \multicolumn{1}{c}{\textbf{Legal}} & \textbf{Cluster} \\ \hline
\multicolumn{1}{|l|}{Driving}  & \multicolumn{1}{c|}{\multirow{3}{*}{ndd}} & \multicolumn{1}{c|}{\multirow{3}{*}{unique}} & \multicolumn{1}{c|}{\multirow{3}{*}{uninterrupted}} & \multicolumn{1}{l|}{A}      & \multicolumn{1}{l|}{yes}   & \multirow{3}{*}{2}                \\ \cline{1-1} \cline{5-6}
\multicolumn{1}{|l|}{Other}    & \multicolumn{1}{c|}{}                     & \multicolumn{1}{c|}{}                        & \multicolumn{1}{c|}{}                               & \multicolumn{1}{l|}{A}      & \multicolumn{1}{l|}{yes}   &                                   \\ \cline{1-1} \cline{5-6}
\multicolumn{1}{|l|}{Break}    & \multicolumn{1}{c|}{}                     & \multicolumn{1}{c|}{}                        & \multicolumn{1}{c|}{}                               & \multicolumn{1}{l|}{DR\_T3} & \multicolumn{1}{l|}{yes}   &                                   \\ \hline
\hline
\multicolumn{7}{|c|}{Description}                                                                                                                                                                                                                           \\ \hline
\multicolumn{7}{|c|}{Legal and standard daily driving formed by a unique and uninterrupted driving sequence}                                                                                                                                               \\ \hline
\end{tabular}%
}
\end{table}


\subsection{Generating Driver Profiles}

Similarly to the working days clustering previously explained, we performed categorisation of drivers based on similar behaviour with the idea of extracting driver profiles. With enough data, we saw that the large amount of activities contained in event logs can be summarised in different types of driving days as described in the previous sections, and such types encode enough information to extract a characterisation of the driver that can be informative for the transport company.

\begin{table}[t]
    \caption{Example input data for extracting driver profiles. Each row encodes how frequently a driver perform one of four types of driving days. Given the uneven number of data of each driver values are expressed as percentages and all rows sums to one.}
    \label{tab:profile}
    \centering
\renewcommand{\arraystretch}{1.2}
\resizebox{\textwidth}{!}{%
\begin{tabular}{|c|llllc|}
\hline
\multirow{2}{*}{\textbf{Driver}} & \multicolumn{5}{c|}{\textbf{Driving day type}} \\ \cline{2-6} 
    & \multicolumn{1}{l|}{Split Sequences Normal Rest} & \multicolumn{1}{l|}{Uninterrupted Sequences Normal Rest} & \multicolumn{1}{l|}{Split Sequences Reduced Rest} & \multicolumn{1}{l|}{Uninterrupted Sequences Reduced Rest} & \dots \\ \hline
    
\multicolumn{1}{|c|}{1} & \multicolumn{1}{r|}{0.5} & \multicolumn{1}{r|}{0.1} 
& \multicolumn{1}{r|}{0.3} & \multicolumn{1}{r|}{0.1} & \\ \hline

\multicolumn{1}{|c|}{2} & \multicolumn{1}{r|}{0.2} & \multicolumn{1}{r|}{0.8} 
& \multicolumn{1}{r|}{0.0} & \multicolumn{1}{r|}{0.0} & \\ \hline

\multicolumn{1}{|c|}{3} & \multicolumn{1}{r|}{0.15} & \multicolumn{1}{r|}{0.5} 
& \multicolumn{1}{r|}{0.3} & \multicolumn{1}{r|}{0.15} & \\ \hline

\end{tabular}
}
\end{table}

We performed the following steps to categorise drivers:
\begin{enumerate}
    \item \textbf{Drop days with infractions}: Tests introducing violations gave us results who grouped drivers by similar ratio of infractions and by day types (e.g., those who tended to excess their break time ended up in the same cluster). However, we opted to not include such information as they did not align with the purposes of the application. Because the context around the infraction cannot be extracted exclusively from the tachograph data (e.g., was the infringement voluntary, due to lack of correct planning or caused by unexpected circumstances on the road?) we believe managers should analyse case by case rather than making decisions without understanding the real motive behind the infraction.
    \item \textbf{Create frequency table}: For each driver its daily logs are processed following the methodology explained in subsections \ref{sec:labelling} and \ref{sec:clustering}, keeping the day type predicted by the clustering model. The training dataset is created counting the frequencies the driver has performed each type of driving day. Due to the fact that in our data the amount of information varies for each driver, these frequencies are transformed into percentages. Ultimately, we obtain a table $D\times C$ as shown in Table \ref{tab:profile}, being $D$ the number of drivers and $C$ the number of different days categories (i.e., the number of clusters discovered in the previous section).
    \item \textbf{Training}: We perform clustering with the resulting table. From a multiple of techniques our best results were obtained with a Gaussian mixture model trained with the Expectation-Maximization algorithm \citep{fraley2002model}.
    
    The deciding factor at choosing the best partition was based exclusively on expert knowledge. Because we were informed that our data was for drivers who performed similar routes on the same country, we looked for a few number of clusters that separated nicely the data.
\end{enumerate}

As a closing point, we would like to note that driver profiles based only on tachograph data could be misleading, as they do not account for the specific routes they perform. We believe a better approach would be to combine the cluster information with route details like distance, type of vehicle or type of cargo, and as a result get a categorisation of the driver given the type of route. That way, decisions based on these profiles will not be biased, and traffic managers could assess their drivers for each particular service. We intend to explore those options in future work.

\section{Experimentation}

\begin{figure*}[!t]
	\centering
    \includegraphics[width=.9\columnwidth]{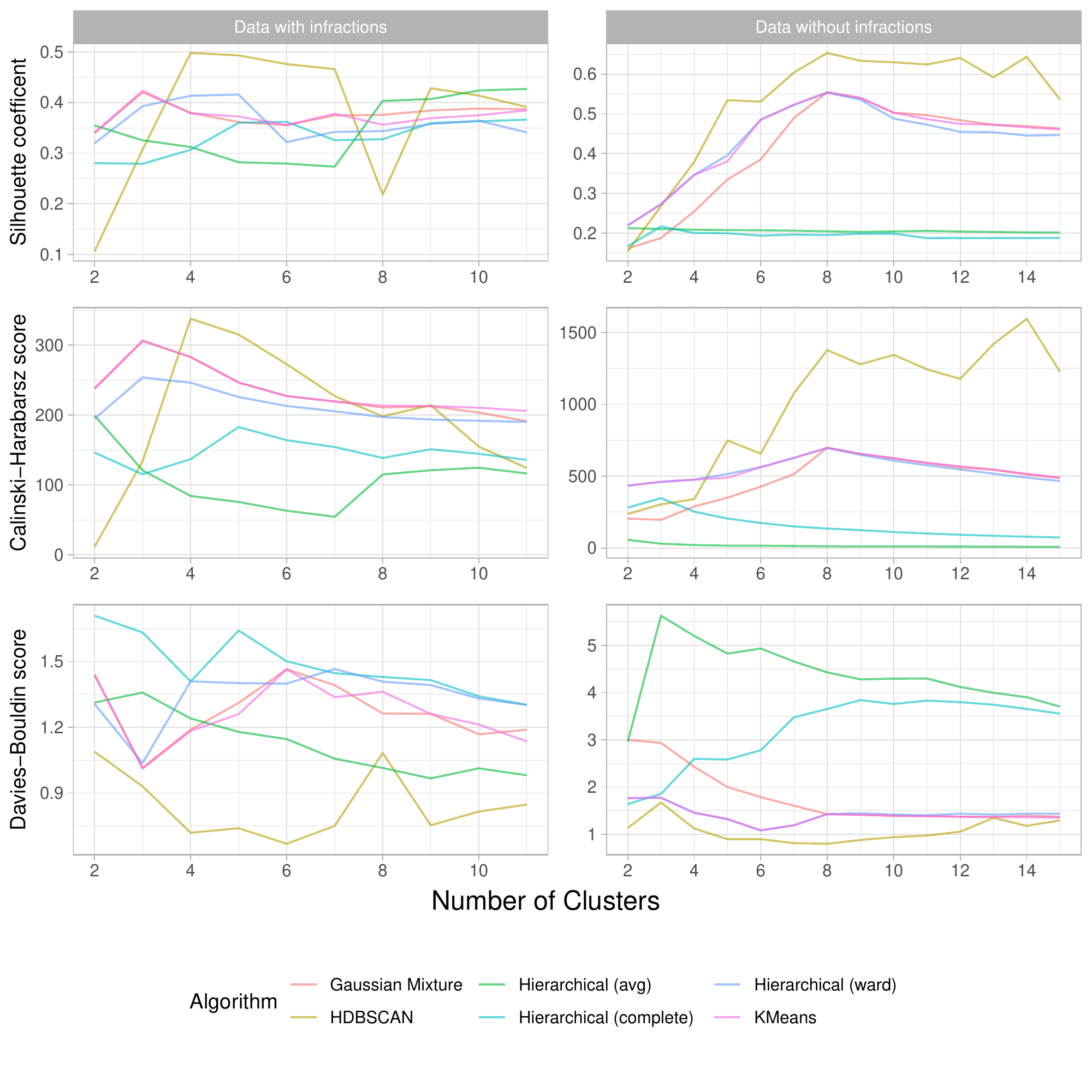}
    \caption{Silhouette Coefficient, Calinski-Harabasz and Davies-Bouldin indexs as a function of the number of clusters for multiple clustering algorithms. Notice that the y-axis scalings differ among the different panels of this figure.}
    \label{fig:scores}
\end{figure*}

We have validated our methodology with an experimentation using real tachograph logs provided by an industrial collaborator. We were provided with a dataset formed by two-weeks-long sequences of activities from 290 different drivers.

Because the architecture is composed of three different components, each one was validated individually. The labelling process was verified against multiple driving sequences selected at random, both legal and illegal, manually verifying that the output was the appropriate under the HoS regulation. For the infringement analysis system multiple tests for each kind of infraction were carried out, confirming that not only the cause, but also the subsequence containing the infraction was detected.

Lastly, due to the fact that the clustering in our problem is an unsupervised task, we experimented with different techniques and hyperparametrization to discover the best possible clusters. The quality of each partition was measured with the Silhouette Coefficient and both Calinski-Harabasz and Davies-Bouldin indexs. The final clustering result was selected between the best performing tests, and after expert inspection of the resulting clusters and centroids.

Figure \ref{fig:scores} shows the performance of multiple algorithms in data with and without infractions, respectively. The algorithms are: Gaussian Mixture models, having each component its own covariance matrix, and controlling the number of mixture components as the number of clusters; HDBSCAN \citep{hdbscan}, a hierarchical clustering model employing density based measures; classical agglomerative clustering using average, complete and ward criteria; and K-Means with cosine similarity as distance metric.

Some insights can be extracted from the graphs. HDBSCAN is without doubt the best algorithm under both subsets of this data, but we believe that the reason relies mostly due to its robustness to noise points. Furthermore, results on data with infringements are, as expected, more variable, as this subset combines multiple types of infractions with driving sequences that can be perfectly legal. The runner-up model is not clear, as results vary greatly with the number of clusters.

For fully legal data we see hierarchical clustering with average and complete linkage method vastly underperforming. The graphs for the rest of techniques take similar shape, mostly agreeing in that 8 clusters seems an appropriate partition for this data. Nevertheless, as the results are intended for human interpretation, it is important to remind that the clusters should be reviewed by an expert whenever possible before setting on a final value.

Finally, we believe is worth mentioning our experimentation with the LDA (Latent Dirichlet Allocation) \citep{lda}. This technique is frequently used in NLP tasks to summarise a document with a set of topics. Due to a small vocabulary size in our data as opposed to an NLP task, most words (\textit{i.e.} driver activities) are present in many different clusters (with the exception of illegal activities), and although the most relevant topics could be ranked and considered as centroids there is no assurance that these topics are understandable (\textit{e.g.} a B\_T2 break only makes sense if followed by a B\_T3 break. The presence of only one of them as a topic does not clarify if the driver completed the sequence or committed an infraction).

\begin{table}[t]
    \centering
    \caption{Clustering results for driver profiles and they interpretation.}
    \label{tab:cluster-result}
\renewcommand{\arraystretch}{1.2}
\resizebox{0.6\textwidth}{!}{%
\begin{tabular}{|c|l|r|}
\hline
\multicolumn{1}{|l|}{\textbf{Cluster}} & \textbf{Interpretation} & \multicolumn{1}{l|}{\textbf{Proportion}} \\ \hline
1 & \thead[l]{No extended days and mostly \\ takes rests uninterrupted} & 8.6\% \\ \hline
2 & \thead[l]{Usually splits rests as much as possible \\ and rarely takes extended days} & 51.8\% \\ \hline
3 & \thead[l]{Neither takes many extended days \\ or splits rests} & 20.2\% \\ \hline
4 & \thead[l]{No clear tendency, \\ driver seems to be flexible} & 14.4\% \\ \hline
5 & \thead[l]{Tends to split rests as much as possible \\ and frequently takes extended days} & 5.0\% \\ \hline
\end{tabular}
}
\end{table}

For our driver clustering experimentation Table \ref{tab:cluster-result} shows 5 resulting clusters and their interpretation after training.  Given that our training data is compromised of mostly event logs of national deliveries in Spain, we can see that more than half of our drives prefer to spent their rests split in two. Nevetheless, as mentioned above, the lack of data about the routes performed in the tachograph hinders the expressivenes, but experts welcome any information that could help them assign the best driver to a service as easily as possible.

The methodology and experimental results are encapsulated in an web application publicly available at \url{https://github.com/IgnacioVellido/Driver-Assistance-System}.

\section{Related Work}

This project is an extension of authors prior work \citep{Fernandez-Olivares_Perez_2020} focused on the recognition and labelling of driver activities under the HoS regulation. The novel contributions provided in this paper go a step forward in our goal of developing an intelligent assistant to drivers and traffic managers, proposing a planning and constraint based analysis of infractions causes and summarisation of driver behaviour with NLP techniques.

Regarding applications concerned with the HoS regulation, many approaches have been developed aimed to solve route planning problems under these rules while minimising transportation costs \citep{gideon, omelianenko2019advanced, goel_legal_2018, goel_exact_2017}. Nonetheless, the authors have not found works that extract insights that can be useful for experts in analysing and understanding driver activities from a legal perspective.

As for driver behaviour modelling from tachograph data, proposals like \citep{zhou2019analysis} employs data mining techniques to categorise truck drivers and analyse dangerous tendencies. Their approach is similar to ours in that clusters are manually studied and labelled. However, PCA for dimensionally reduction and DBSCAN for clustering are directly used instead due to the fact that their data does not contain categorical variables.

Lastly, word embedding techniques like Paragraph Vector models has been previously applied in non textual data like web user activities \citep{tagami2015modeling} and server logs \citep{mimura2018leaving} as a way to transform sequential data of variable length into dimensionally fixed data. Similarly, although oriented to process mining applications, the \textit{trace2vec} model proposed in \citep{trace2vec} uses embedding techniques for discovery, monitoring and clustering of sequences of activities. Nevertheless, up to the author's knowledge there have not been prior research with tachograph logs.


\section{Conclusion}

We have presented a novel planning application that brings the worlds of Data Analytics, IoT and Automated Planning and Scheduling together. The approach provides support to experts on the task of interpreting what drivers are or have been doing by recognising and summarising their activity recorded in an event log.

Using as a basis our prior work in driver activity recognition, the main contributions exposed in this paper are an infringement analysis process with a planning and constraint based approach, the summarisation of temporal activity logs using word embeddings and clustering, and the creation of driver profiles based on such summaries. The overall system provides a human readable summary of the driver behaviour under the HoS regulation while explaining infractions and the root cause.

Regarding future work, it is worth noting that the main interest and the ultimate goal of the company is to build an intelligent assistant to provide decision support services to both drivers and companies decision makers. This is a research direction aligned with the concept of assistive interaction \citep{freedman2017integration}, that advocates for the integration of plan recognition and planning. In this way, the recognition of driver's intent is a previous stage needed to respond with a generated plan adapted to the currently recognised task.

For our next steps we intent to enrich the driver profiling model adding non-tachograph data about the transport service, like type of vehicle and cargo. Additionally, we are focused on integrating descriptive support to the assistant, being able to suggest drivers plans of actions in compliance with the HoS regulation and considering preference patterns extracted from previous personal behaviour.


\bibliographystyle{kluwer}
\bibliography{bibliography}

\end{document}